%% file: mainConferenceCIARP.tex
\documentclass[runningheads]{llncs}
\usepackage{graphicx}
\usepackage{cellspace}

\begin{document}
\title{Spatiotemporal CNNs for Pornography Detection in Videos}
%
%

\author{Murilo Varges da Silva\inst{1,2} \and Aparecido Nilceu Marana\inst{3}}

%
%
\institute{UFSCar - Federal University of Sao Carlos, Sao Carlos, SP, Brazil \and
IFSP - Federal Institute of Education of Sao Paulo, Birigui, SP, Brazil \and
UNESP - Sao Paulo State University, Bauru, SP, Brazil
\email{murilo.varges@ifsp.edu.br,nilceu.marana@unesp.br }
}

\maketitle              

\input{0_abstract.tex}
\input{1_introduction.tex}
\input{2_related_works.tex}
\input{3_methods.tex}
\input{4_experiments_results.tex}
\input{5_conclusion.tex}
\input{6_acknowledgment.tex}

\bibliographystyle{splncs04}
\bibliography{references/ref}

\end{document}

%% file: 0_abstract.tex
\begin{abstract}

With the increasing use of social networks and mobile devices, the number of videos posted on the Internet is growing exponentially. Among the inappropriate contents published on the Internet, pornography is one of the most worrying as it can be accessed by teens and children. Two spatiotemporal CNNs, VGG-C3D CNN and ResNet R(2+1)D CNN, were assessed for pornography detection in videos in the present study. Experimental results using the Pornography-800 dataset showed that these spatiotemporal CNNs performed better than some state-of-the art methods based on bag of visual words and are competitive with other CNN-based approaches, reaching accuracy of $95.1\%$.  

\keywords{Pornography detection, Spatiotemporal CNN, 3D CNN, Video classification.}

\end{abstract}

%% file: 1_introduction.tex
\section{Introduction}
\label{sec:introduction}

With the increasing use of social networks and mobile devices, the number of videos posted on the Internet is growing exponentially. Thus, the recognition of videos with unwanted content (e.g. pornography, violence, scenes with blood, etc.) becomes essential. Pornography is probably the type of unwanted content that causes most problems because it is inappropriate for some ages, inappropriate for some environments (e.g. public places, schools, workplace), and unwanted by some people who do not like to be exposed to this material. Another important aspect is that some of this content might be prohibited by law from being produced and disseminated, as in the case of child pornography, which is considered a crime in several countries. 



%% file: 2_related_works.tex

Video understanding is a challenging task in the fields of Computer Vision and Pattern Recognition and has been studied for many decades. Much of the research developed in these areas are focused on creating spatiotemporal descriptors for video understanding. The most relevant studies that deal with hand-crafted feature extraction from videos include those based on spatiotemporal interest points: STIPs (HARRIS3D)\cite{Laptev2003}, SIFT-3D\cite{Scovanner2007}, HOG3D\cite{klaser2008}, MBH\cite{Dalal2006} and Cuboids\cite{Dollar2005}. These efforts are based on 2D image descriptors and use different encoding schemes based in pyramids and histograms. In addition to these, another very important state-of-the-art method is the improved Dense Trajectories (iDT)\cite{Wang2013}, which presents good performance in tasks related to video understanding.

The great development of methodologies that use deep learning in still-image recognition tasks, driven by the development of the AlexNet network\cite{Krizhevsky2012}, increased the interest in research using deep learning techniques applied to videos. Some approaches proposed to apply trained CNN in images to extract features from individual video frames and then fuse these features into a descriptor with fixed size using pooling and high-dimensional encoding. Another alternative is to use 3D spatiotemporal CNNs. Ji et al.\cite{Yang2015}, for instance, proposed 3D CNN spatiotemporal convolutions to recognize human actions in videos.  
Simonyan and Zisserman\cite{Simonyan2014-Two-Stream} introduced a new influential two-stream framework approach based on CNNs, in which deep motion features extracted from the optical flow are fused with traditional CNN activations computed from RGB input. In \cite{Perez2017} the authors use this two-stream framework approach based on CNNs to detect pornography in videos.

Two spatiotemporal-based CNNs proposed in literature, VGG-C3D CNN \cite{Tran2015} and ResNet R(2+1)D CNN \cite{Tran2017}, were used in the present study for pornography detection in videos. To the best of our knowledge, this is the first study to use 3D CNN to detect pornography in videos.

%% file: 3_methods.tex


\section{VGG C3D CNN}
\label{sec:vgg-c3d}

\begin{figure*}[htbp]
\centerline{\includegraphics[scale=0.18]{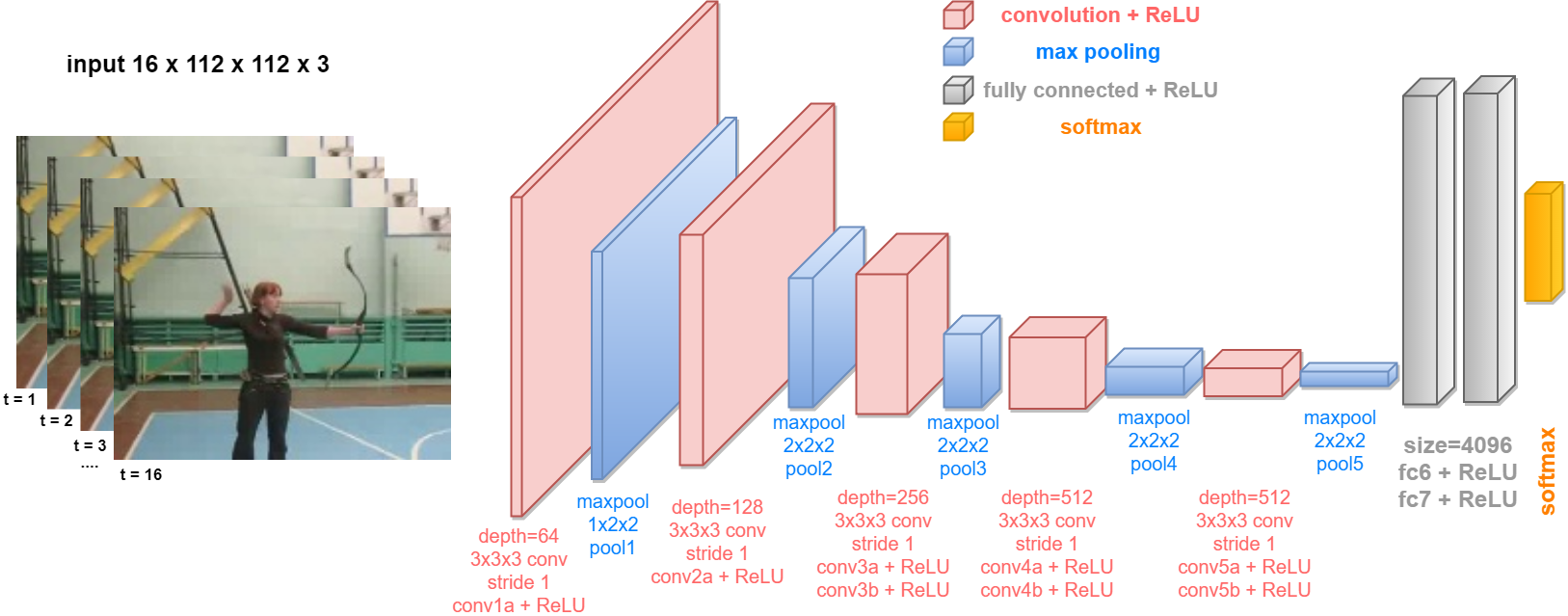}}
\caption{VGG-C3D architecture based in VGG-11 \cite{Simonyan2014-VGG} proposed by \cite{Tran2015}. 
}
\label{fig3dcnn}
\end{figure*}

In \cite{Tran2015} the authors realized that a homogeneous setting with convolution kernels of $3\times3\times3$ is the best option for 3D CNNs. This is similar to the 2D CNNs proposed in \cite{Simonyan2014-VGG}, which are also known as VGG.
By using a dataset with a huge amount of data, it is possible to train a 3D CNN with $3\times3\times3$ kernel as deep as possible, due to the amount of memory available in current GPUs. The authors designed the 3D CNN to have 8 convolution layers, 5 pooling layers, followed by two fully connected layers, and a softmax output layer. Fig. \ref{fig3dcnn} shows the 3D Spatiotemporal CNN proposed in \cite{Tran2015} (called VGG-C3D in this paper). The 3D convolution filters of VGG-C3D are of dimension $3\times3\times3$ with stride $1\times1\times1$. In turn, the 3D pooling layers are $2\times2\times2$ with stride also of $2\times2\times2$, except for pool1 which presents kernel size of $1\times2\times2$ and stride $1\times2\times2$ with the intention of preserving the temporal information at the early phase. Each fully connected layer has 4,096 output units.

The model provided by the authors \cite{Tran2015}, which was trained on the Sports-1M dataset in train split, was used in the present study. Sports-1M was created by Google Research and Stanford Computer Science Department and contains 1,133,158 videos of 487 sports classes. Since Sports-1M has many long videos, five 2-seconds long clips were randomly extracted from every training video. The clips were then resized to have a frame size of $128\times171$. During the training phase, the clips were randomly cropped into $16\times112\times112$ crops for spatial and temporal jittering, and horizontally flipped with 50\% probability. The training was done by Stochastic Gradient Descent (SGD) with a minibatch size of 30 examples. The initial learning rate was of 0.003 and was divided by 2 every 150K iterations. The optimization was stopped at 1.9M iterations (about 13 epochs).

After training, the VGG-C3D may be used as feature extractor. In order to extract features, a video needs to be split into clips with 16 frames in length. For the present study, clips with an 8-frame overlap between two consecutive clips were used. After that, the clips were submitted to the VGG-C3D to extract \textbf{fc6} activations. Each video may have an arbitrary number of clips, so to generate only one descriptor for each video the \textbf{fc6} activations were averaged to form a 4,096-sized descriptor, followed by L2-normalization.

\begin{figure*}[htbp]
\begin{minipage}[b]{0.45\linewidth}
  \centering
  \centerline{\includegraphics[height=0.7\linewidth]{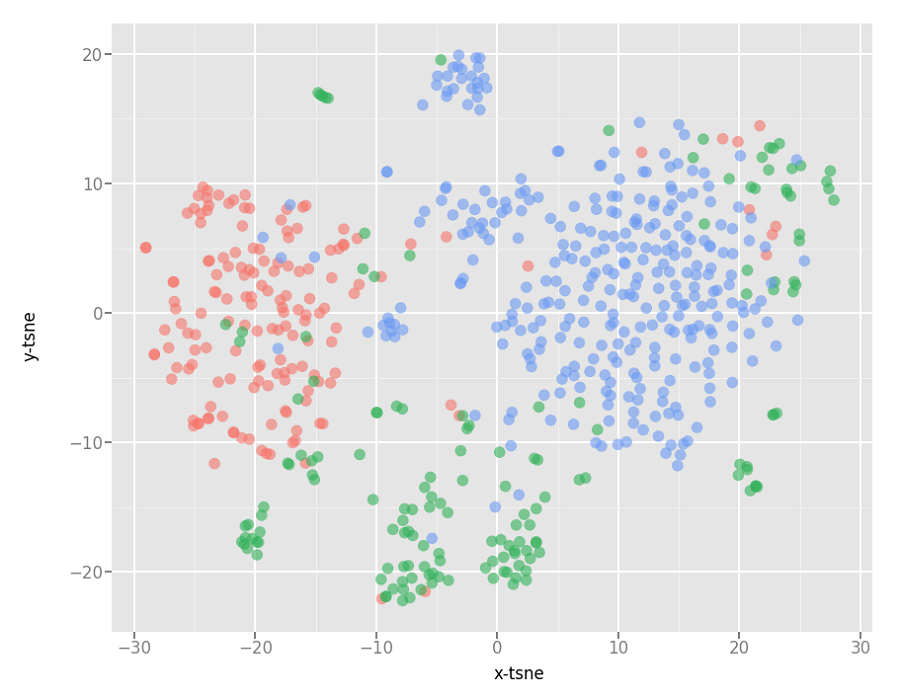}}
  \centerline{(a)}\medskip
\end{minipage}
\begin{minipage}[b]{0.45\linewidth}
  \centering
  \centerline{\includegraphics[height=0.7\linewidth]{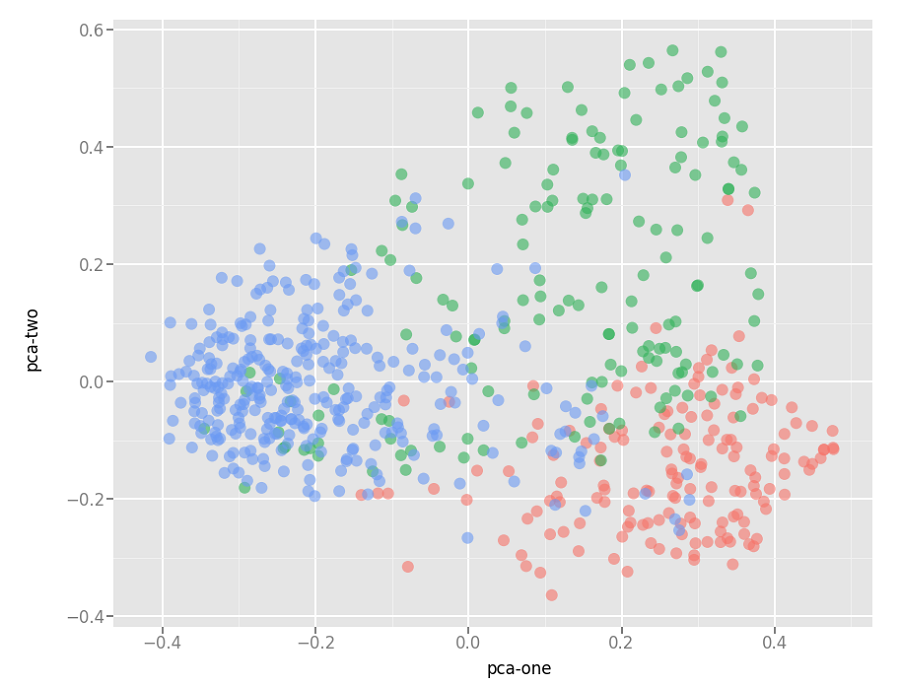}}
  \centerline{(b)}\medskip
\end{minipage}
\caption{Feature embedding visualizations of VGG-C3D on samples from Pornography-800 dataset: pornography (blue), easy non-pornography  (red) and difficult non-pornography  (green). (a) Using t-SNE and (b) Using PCA.}
\label{fig:features}
\end{figure*}

To evaluate the VGG-C3D features extracted from the Pornography-800 dataset \cite{AVILA2013}, \textbf{fc6} features were extracted from all clips and then projected to 2D spacing using the t-SNE \cite{vanDerMaaten2008-tSNE} (Fig. \ref{fig:features}a) and PCA (Fig. \ref{fig:features}b). It is worth noting that no fine-tuning was conducted to verify if the model showed good generalization capability across the datasets. Fig. \ref{fig:features} illustrates that the VGG-C3D features are semantically separable, although samples of pornography and difficult non-pornography presented some overlapping.\textbf{}


\section{ResNet R(2+1)D CNN}
\label{sec:r+1d}

Recent studies have indicated that replacing 3D convolutions by two operations, a 2D spatial convolution and a 1D temporal convolution, can improve the efficiency of 3D CNN models. In \cite{Tran2017}, the authors designed a new spatiotemporal convolutional block, R(2+1)D, that explicitly factorizes 3D convolution into two separate and successive operations, a 2D spatial convolution and a 1D temporal convolution. Using this architecture, we can add nonlinear rectification like ReLU between 2D and 1D convolution. This would double the number of nonlinearities compared to a 3D CNN, but with the same number of parameters to optimize, allowing the model to represent more complex functions. Moreover, the decomposition into two convolutions makes the optimization process easier, producing in practice less training loss and less test loss. 

Another method proposed by \cite{Saining2017} showed that replacing 3D convolutions with spatiotemporal-separable 3D convolutions makes the model 1.5x more computationally efficient (in terms of FLOPS) than 3D convolutions.
 
Experiments performed in \cite{Tran2017} demonstrated that ResNets adopting homogeneous (2+1)D blocks in all layers, achieved state-of-the-art performance on both Kinetics and Sports-1M datasets.

\begin{figure}[htbp]
\begin{minipage}[b]{0.43\linewidth}
  \centering
  \centerline{\includegraphics[height=0.45\linewidth]{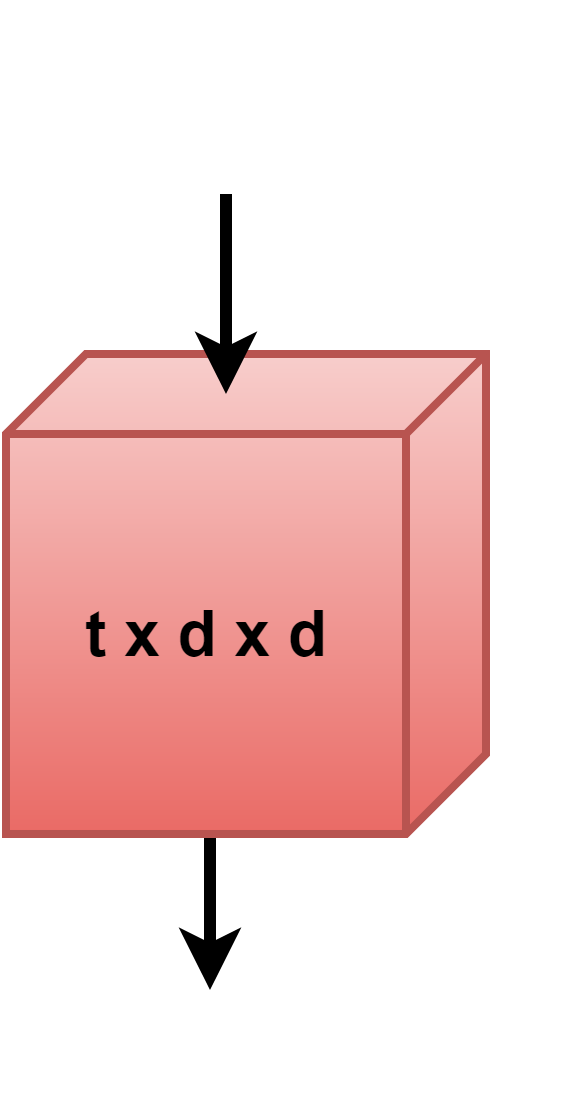}}
  \centerline{(a)}\medskip
\end{minipage}
\begin{minipage}[b]{0.43\linewidth}
  \centering
  \centerline{\includegraphics[height=0.45\linewidth]{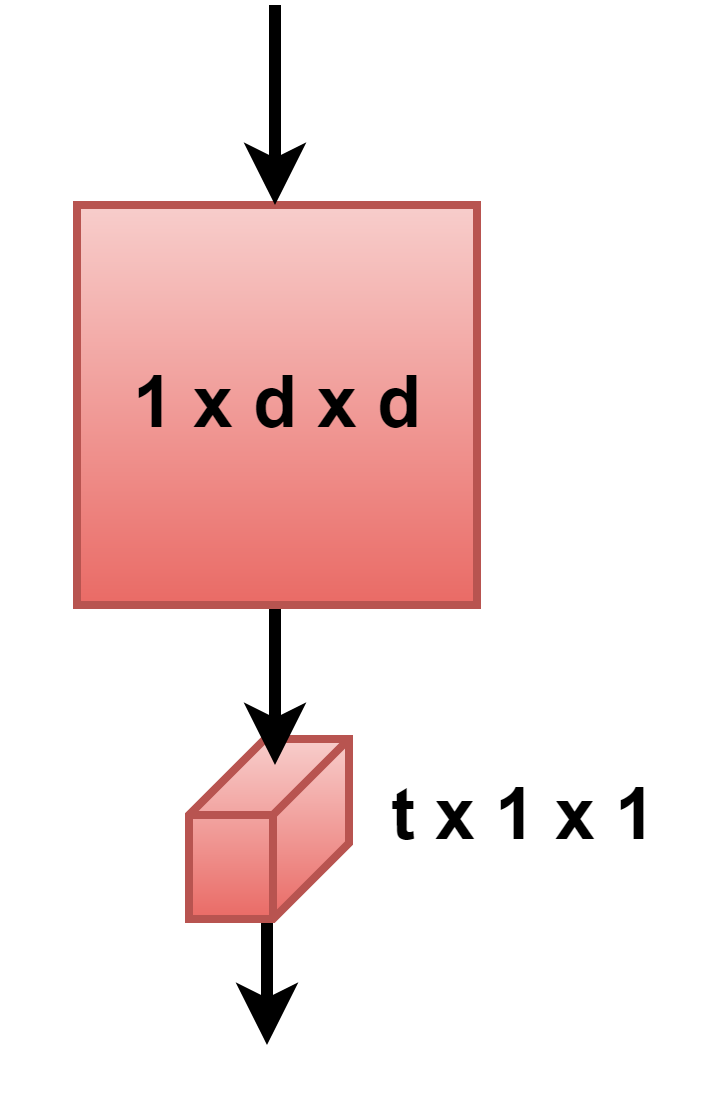}}
  \centerline{(b)}\medskip
\end{minipage}
\caption{3D convolution \textit{vs} (2+1)D convolution. 
(a) Full 3D convolution using a filter of the size $t\times d \times d$, where $t$ denotes the temporal extent and $d$ is the spatial width and height. (b) A (2+1)D convolutional block, where a spatial 2D convolution is followed by a temporal 1D convolution. Adapted from \cite{Tran2017}.}
\label{fig:r2+1d}
\end{figure}

Spatiotemporal decomposition can be applied to any 3D convolutional layer. An illustration of this decomposition is given in Fig. \ref{fig:r2+1d} for the simplified setting, where the input tensor contains a single channel.

The architecture proposed in \cite{Tran2017} was applied in the present study. This relatively simple structure was based on deep residual networks, which have shown good performance. Table \ref{tab-resnet} presents the architecture details of R(2+1)D. 

\begin{table}[htb]
\caption{R(2+1)D architecture\cite{Tran2017} used in the present study. 
}
\begin{center}
\scalebox{0.7}{
\setlength\cellspacetoplimit{1pt}
\setlength\cellspacebottomlimit{1pt}
\begin{tabular}{Sc|Sc|Sc}
\hline
\textbf{layer name} & \textbf{output size} &\textbf{34-layer}\\
\hline
conv1 & $L\times5\times56$  &$3\times7\times7,64$, stride $1\times2\times2$ \\
\hline
conv2 & $L\times56\times56$  & $\left[ \begin{array}{cc} 3\times3\times3,64  \\ 3\times3\times3,64 \end{array}\right] \times 3$  \\
\hline
conv3 & $\frac{L}{2}\times28\times28$  &  $\left[ \begin{array}{cc} 3\times3\times3,128  \\ 3\times3\times3,128 \end{array}\right] \times 4$  \\
\hline
conv4 & $\frac{L}{4}\times14\times14$  &  $\left[ \begin{array}{cc} 3\times3\times3,256  \\ 3\times3\times3,256 \end{array}\right] \times 6$  \\
\hline
conv5 & $\frac{L}{8}\times7\times7$  &  $\left[ \begin{array}{cc} 3\times3\times3,512  \\ 3\times3\times3,512 \end{array}\right] \times 3$  \\
\hline
 & $1\times1\times1$ & spatiotemporal pooling, fc layer with softmax  \\
\hline
\end{tabular}}
\renewcommand{\arraystretch}{1}
\label{tab-resnet}
\end{center}
\end{table}

Experiments were conducted using a model that had been pre-trained on Kinetics dataset. A transfer learning technique was applied to fine-tune the model on the Pornography-800 dataset. The R(2+1)D network used had 34 layers and videos frames were resized to $128\times171$, with each clip generated by randomly cropping $112\times112$ windows. A total of 32 consecutive frames were randomly sampled from each video applying temporal jittering during the process of fine-tuning.

Although Pornography-800 has only about 640 training videos in each split, epoch size was set at 2,560 for temporal jittering considering 4 clips for each training video per epoch. This setup was chosen to optimize the training time since the videos have different sizes.

Batch normalization was applied to all convolutional layers and mini-batch size was set to 4 clips due to GPU memory limitations. The initial learning rate was set to 0.0001 and divided by 10 every 2 epochs, while the process of fine-tuning was conducted in 8 epochs. In the classification phase, the videos were split into 32-frame long clips. ResNet R(2+1)D CNN was used on clips with 16 frames that overlap between two consecutive clips to extract features and for softmax classification. Each video can have an arbitrary number of clips, so average pooling on softmax probabilities was conducted to aggregate predictions over clips to obtain video-level prediction.

%% file: 4_experiments_results.tex
\section{Experiments and Results}
\label{sec:experiments}



\begin{figure}[htbp]
\centerline{\includegraphics[scale=0.27]{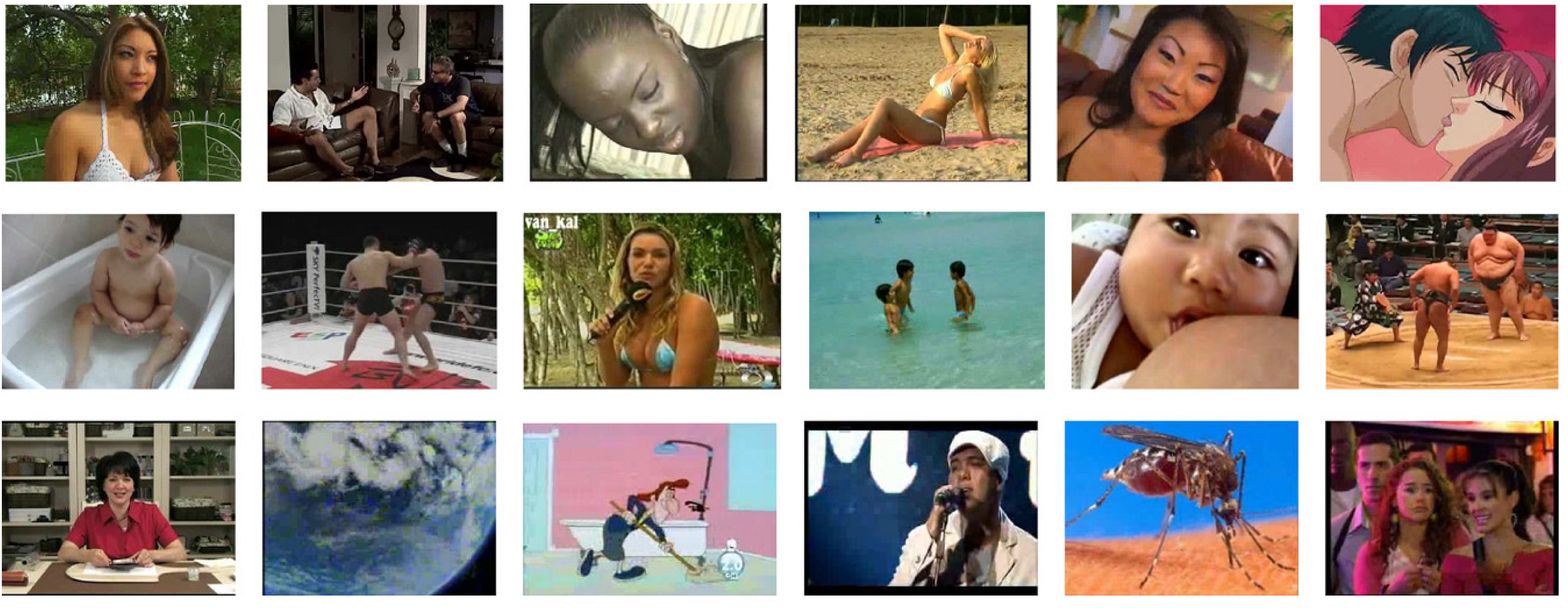}}
\caption{Pornography-800 dataset\cite{AVILA2013}. Top row: pornographic videos. Middle row: challenging cases of non-pornographic videos. Bottom row: easy cases of non-pornographic videos.}
\label{fignpdi}
\end{figure}

The Pornography-800 dataset \cite{AVILA2013} was chosen to evaluate the 3D CNNs used in the present study. This dataset contains 800 videos, representing a total of 80 hours, which encompass 400 pornography videos and 400 non-pornography videos.
Fig. \ref{fignpdi} shows some selected frames from a small sample of this dataset, illustrating the diversity and challenges posed.

The VGG-C3D network evaluated in the present study was developed using Caffe\cite{jia2014caffe} and the ResNet R(2+1)D architecture was developed using Caffe2\footnote{ https://caffe2.ai/}. All experiments were run on a computer with an Intel Xeon E5-2630 v3 2.40GHz processor, 32 GB RAM and a NVIDIA Titan XP GPU with 12 GB of memory.
The results presented are the mean value obtained from the 5 splits of the Pornography-800 dataset using 5-fold-cross-validation protocol (640 videos in training set and 160 in the test set on each fold, which is the same protocol proposed in \cite{AVILA2013}).

Table \ref{tab:results} shows the accuracy of both approaches: VGG-C3D with a Linear SVM classifier and ResNet R(2+1)D CNN with softmax classifier. The VGG-C3D architecture with a linear SVM classifier achieved a better performance, with accuracy of 95.1\%, while with the ResNet R(2+1)D architecture using the softmax classifier achieved accuracy of 91.8\%.

\begin{table}[htbp]
\caption{Results achieved by VGG-C3D (Section \ref{sec:vgg-c3d}) and ResNet R(2+1)D (Section \ref{sec:r+1d}) on the Pornography-800 dataset.}
\begin{center}
\scalebox{0.8}{
\setlength\cellspacetoplimit{1pt}
\setlength\cellspacebottomlimit{1pt}
\begin{tabular}{Sc|Sc}
\hline
\textbf{Approach} & \textbf{Accuracy(\%)} \\
\hline 
VGG-C3D + Linear SVM & \textbf{95.1 $\pm$ 1.7} \\
ResNet R(2+1)D CNN + Softmax & 91.8 $\pm$ 2.1 \\
\hline
\end{tabular}}
\label{tab:results}
\end{center}
\end{table}

Table \ref{tab:comparison} presents some  results  reported in the literature obtained applying other methods to the Pornography-800 database. As observed in Tables \ref{tab:results} and \ref{tab:comparison} the CNN-based methods outperform all methods based on bag of visual words.  

\begin{table}[htbp]
\caption{Results obtained by state-of-the-art methods for pornography video detection on the Pornography-800 dataset. 
}
\begin{center}
\scalebox{0.8}{
\setlength\cellspacetoplimit{1pt}
\setlength\cellspacebottomlimit{1pt}

\begin{tabular}{Sc|Sc|Sc|Sc}
\hline
\textbf{Approach} & \textbf{Reference} & \textbf{Year} & \textbf{Accuracy(\%)} \\
\hline
BoVW-Based  & Avila et al.\cite{Avila2011} & 2011 & 87.1\% $\pm$ 2.0  \\
            & Valle et al.\cite{Valle2011} & 2011 & 91.9\% $\pm$ NA \\
            & Souza et al.\cite{Souza2012} & 2012 & 91.0\% $\pm$ NA \\
            & Avila et al.\cite{AVILA2013} & 2013 & 89.5\% $\pm$ 1.0 \\
            & Caetano et al.\cite{caetano2014} & 2014 & 90.9\% $\pm$ 1.0 \\
            & Caetano et al.\cite{Caetano2016} & 2016 & 92.4\% $\pm$ 2.0 \\
            & Moreira et al. \cite{MOREIRA2016} & 2016 & 95.0 \% $\pm$ 1.3 \\
\hline
2D CNN RGB  & Moustafa\cite{Moustafa2015} & 2015 & 94.1\% $\pm$ 2.0 \\
            &  Perez et al.\cite{Perez2017} & 2017 & 97.0\% $\pm$ 2.0 \\
\hline
2D CNN OF   &  Perez et al.\cite{Perez2017} & 2017 &95.8\% $\pm$ 2.0 \\
\hline
Two Stream CNN & Perez et al.\cite{Perez2017} & 2017 &\textbf{97.9\% $\pm$ 1.5} \\

\hline
\end{tabular}}
\label{tab:comparison}
\end{center}
\end{table}

%% file: 5_conclusion.tex
\section{Conclusion and Future Work}
\label{sec:conclusion}

The experimental results obtained in the present study on the Pornography-800 dataset showed that the spatiotemporal CNNs adopted (VGG-C3D and ResNet R(2+1)D) performed better than all methods based on bag of visual words compared. Moreover, these spatiotemporal CNNs were competitive with other CNN-based approaches observed, reaching accuracy of $95.1\%$. 

With recent creation and availability of large video databases, along with the evolution of GPUs, we believe that 3D CNNs will be able to achieve a state-of-the-art level in video understanding tasks, similar to what happened with the launch of AlexNet. The proof of this is that a 3D CNN (I3D)\cite{Carreira2017-I3D} has recently reached the best result in the Kinetics database.

Future research is expected to: apply VGG-C3D and ResNet R(2+1)D CNNs to larger databases; fuse VGG-C3D features with the iDT; to evaluate the behavior of the 3D CNNs using the Optical Flow and the fusion with the RGB; and use the ResNet R(2+1)D CNN as feature extractor.



%% file: 6_acknowledgment.tex
\section*{Acknowledgments}
\label{sec:acknowledgments}

We thank NVIDIA Corporation for the donation of the GPU used in this study. This study was financed in part by CAPES - Brazil (Finance Code 001).